\let\oldvec\vec
\documentclass{llncs}
\let\vec\oldvec

\usepackage{algorithm}
\usepackage{algpseudocode}
\usepackage{amsfonts}
\usepackage{amsmath}
\usepackage{amssymb}
\usepackage{color}
\usepackage{enumerate}  
\usepackage{enumitem}
\usepackage{fancybox}
\usepackage{graphicx}
\usepackage{hyperref}
\usepackage{longtable}
\usepackage{makeidx}  
\usepackage{outlines}
\usepackage{rotating}
\usepackage{subcaption}
\usepackage{tabularx}
\usepackage{url}
\usepackage{xcolor}
\usepackage{xspace}

\usepackage{cleveref}
\crefname{equation}{equation}{equations}
\Crefname{equation}{Equation}{Equations}
\crefrangelabelformat{equation}{(#3#1#4--#5#2#6)}
\crefmultiformat{equation}{equations (#2#1#3}{, #2#1#3)}{#2#1#3}{#2#1#3}
\Crefmultiformat{equation}{Equations (#2#1#3}{, #2#1#3)}{#2#1#3}{#2#1#3}

\begin{document}

\frontmatter          

\pagestyle{headings}  

\mainmatter              

\title{A Linear-Time Variational Integrator \\ for Multibody Systems}
\author{Jeongseok Lee\inst{1} \and C. Karen Liu\inst{2} \and Frank C. Park\inst{3} \and Siddhartha S. Srinivasa\inst{1}}
\institute{Carnegie Mellon University, Pittsburgh, PA, USA 15213\\
\email{\{jeongsel, ss5\}@andrew.cmu.edu} \and
Georgia Institute of Technology, Atlanta, GA, USA 30332\\
\email{karenliu@cc.gatech.edu} \and
Seoul National University, Seoul 151-742, Korea\\
\email{fcp@snu.ac.kr}}



\maketitle

%
%
%

\newcommand\todo[1]{\textcolor{red}{TODO: #1}}
\newcommand\sidd[1]{\textcolor{blue}{Sidd: #1}}
\newcommand\karen[1]{\textcolor{magenta}{Karen: #1}}
\newcommand\frank[1]{\textcolor{purple}{Frank: #1}}
\newcommand\js[1]{\textcolor{orange}{JS: #1}}






\newcommand {\SOthree}	{\mathsf{SO}(3)\xspace}
\newcommand {\SOtwo}	{\mathsf{SO}(2)\xspace}
\newcommand {\SOn}		{\mathsf{SO}(n)\xspace}
\newcommand {\sothree}	{\mathfrak{so}(3)\xspace}
\newcommand {\sotwo}	{\mathfrak{so}(2)\xspace}
\newcommand {\son}		{\mathfrak{so}(n)\xspace}

\newcommand {\SEthree}	{\mathsf{SE}(3)\xspace}
\newcommand {\SEtwo}	{\mathsf{SE}(2)\xspace}
\newcommand {\SEn}		{\mathsf{SE}(n)\xspace}
\newcommand {\sethree}	{\mathfrak{se}(3)\xspace}
\newcommand {\dsethree}	{\mathfrak{se^*}(3)\xspace}
\newcommand {\setwo}	{\mathfrak{se}(2)\xspace}
\newcommand {\sen}		{\mathfrak{se}(n)\xspace}

\newcommand {\f}[1]    	{f_{#1}}
\newcommand {\T}[1]     {T_{#1}}
\newcommand {\iT}[1]	{T_{#1}^{-1}}
\newcommand {\dT}[1]	{\dot{T}_{#1}}
\newcommand {\Tlam}[1]	{T_{\lambda(#1),#1}}
\newcommand {\iTlam}[1]	{T_{\lambda(#1),#1}^{-1}}

\newcommand \Ad[1]		{\text{Ad}_{#1}}
\newcommand \dAd[1]		{\text{Ad}^{*}_{#1}}
\newcommand \Me[1]      {M_{#1}e^{S_{#1}q_{#1}}}
\newcommand \MeSq[1]	{M_{#1}e^{S_{#1}q_{#1}}}
\newcommand \MeAq[1]	{M_{#1}e^{A_{#1}q_{#1}}}
\newcommand \MeBq[1]	{M_{#1}e^{B_{#1}q_{#1}}}
\newcommand \MeSt[1]	{M_{#1}e^{S_{#1}\theta_{#1}}}
\newcommand \MeAt[1]	{M_{#1}e^{A_{#1}\theta_{#1}}}
\newcommand \MeBt[1]	{M_{#1}e^{B_{#1}\theta_{#1}}}
\newcommand \Adf[1]		{\text{Ad}_{f_{#1}}}
\newcommand \Adif[1]	{\text{Ad}_{f_{#1}^{-1}}}
\newcommand \dAdf[1]    {\text{Ad}^*_{f_{#1}}}
\newcommand \dAdif[1]	{\text{Ad}^*_{f_{#1}^{-1}}}
\newcommand \AdT[1]		{\text{Ad}_{T_{#1}}}
\newcommand \AdiT[1]	{\text{Ad}_{T_{#1}^{-1}}}
\newcommand \dAdT[1]    {\text{Ad}^*_{T_{#1}}}
\newcommand \dAdiT[1]	{\text{Ad}^*_{T_{#1}^{-1}}}
\newcommand \adv[1]     {\text{ad}_{V_{#1}}}
\newcommand \dadv[1]    {\text{ad}^*_{V_{#1}}}
\newcommand \adV[1]     {\text{ad}_{V_{#1}}}
\newcommand \dadV[1]    {\text{ad}^*_{V_{#1}}}
\newcommand \ad[1]      {\text{ad}_{#1}}
\newcommand \dad[1]     {\text{ad}^*_{#1}}
\newcommand \dq[1]      {\dot{q}_{#1}}
\newcommand \ddq[1]  	{\ddot{q}_{#1}}

\newcommand \V[1]       {V_{#1}}
\newcommand \Vi			{V_i}
\newcommand \Vlam[1]	{V_{\lambda(#1)}}
\newcommand \dV[1]      {\dot{V}_{#1}}
\newcommand \dVlam[1]	{\dot{V}_{\lambda(#1)}}

\newcommand \pq[1]		{\frac{\partial {#1}}{\partial q}}
\newcommand \pqi[1]		{\frac{\partial {#1}}{\partial q_i}}
\newcommand \pqj[1]		{\frac{\partial {#1}}{\partial q_j}}
\newcommand \pdq[1]		{\frac{\partial {#1}}{\partial \dot q_i}}
\newcommand \pdqi[1]	{\frac{\partial {#1}}{\partial \dot q_i}}
\newcommand \pdqj[1]	{\frac{\partial {#1}}{\partial \dot q_j}}
\newcommand \pqq[1]		{\frac{\partial^2 {#1}}{\partial q_i\partial q_j}}
\newcommand \pdqq[1]	{\frac{\partial^2 {#1}}{\partial \dot p_i\partial q_j}}
\newcommand \pqdq[1]	{\frac{\partial^2 {#1}}{\partial q_i\partial \dot q_j}}
\newcommand \pdqdq[1]	{\frac{\partial^2 {#1}}{\partial \dot q_i\partial \dot q_j}}

\newcommand \fpp[2]     {\frac{\partial {#1}}{\partial {#2}}}
\newcommand \fppp[3]    {\frac{\partial^2 {#1}}{\partial {#2} \partial {#3}}}
\newcommand \pp[1]      {\frac{\partial {#1}}{\partial p}}
\newcommand \ppi[1]     {\frac{\partial {#1}}{\partial p_i}}
\newcommand \ppj[1]		{\frac{\partial {#1}}{\partial p_j}}
\newcommand \ppk[1]		{\frac{\partial {#1}}{\partial p_k}}
\newcommand \ppl[1]		{\frac{\partial {#1}}{\partial p_l}}
\newcommand \ppm[1]     {\frac{\partial {#1}}{\partial p_m}}
\newcommand \ppn[1]     {\frac{\partial {#1}}{\partial p_n}}
\newcommand \ppij[1]    {\frac{\partial^2 {#1}}{\partial p_i\partial p_j}}
\newcommand \ppmn[1]    {\frac{\partial^2 {#1}}{\partial p_m\partial p_n}}
\newcommand \ppkl[1]    {\frac{\partial^2 {#1}}{\partial p_k\partial p_l}}
\newcommand \pfpq[1]    {\frac{\partial f}{\partial q_{#1}}}

\newcommand \fdd[2]     {\frac{d {#1}}{d {#2}}}
\newcommand \fddd[3]    {\frac{d^2 {#1}}{d {#2} d {#3}}}

\newcommand {\p}        	{\partial}
\newcommand {\n}        	{\nabla}
\newcommand \calA			{\mathcal{A}}
\newcommand \calD			{\mathcal{D}}
\newcommand \calM			{\mathcal{M}}
\newcommand \calG			{\mathcal{G}}
\newcommand \calC			{\mathcal{C}}
\newcommand \calS			{\mathcal{S}}
\newcommand \calL			{\mathcal{L}}
\newcommand \frakg			{\mathfrak{g}}
\newcommand \frakh			{\mathfrak{h}}
\newcommand \frakX			{\mathfrak{X}}

\newcommand \dtau			{\textit{d}\tau}
\newcommand \itau			{\tau^{-1}}
\newcommand \idtau			{{\textit{d}\tau}^{-1}}

\newcommand \dexp			{\text{\textit{d}exp}}
\newcommand \iexp			{\text{exp}^{-1}}
\newcommand \idexp			{\text{\textit{d}exp}^{-1}}

\newcommand \dcay			{\text{\textit{d}cay}}
\newcommand \icay			{\text{cay}^{-1}}
\newcommand \idcay			{\text{\textit{d}cay}^{-1}}

\newcommand \dlog			{\text{\textit{d}log}}
\newcommand \ilog			{\text{log}^{-1}}

\let\ret\tau
\let\dret\dtau
\let\iret\itau
\let\idret\idtau


\newcommand \SI             {G}

\newcommand \aI[1]			{\hat{\mathcal{I}}_{#1}}
\newcommand \I[1]			{\mathcal{I}_{#1}}
\newcommand \aB[1]			{\hat{\mathcal{B}}_{#1}}

\newcommand \shat[3]        {\hat{#1}_{#2}^{#3}}

\newcommand \pose[2]        {T_{#1}^{#2}}

\newcommand{\bomega}			{\ensuremath{\mathbf{\omega}}}
\newcommand{\blambda}		{\ensuremath{\mathbf{\lambda}}}

\newcommand {\defeq}			{\ensuremath{\triangleq}}
\newcommand {\trans}			{\ensuremath{^{\mbox{\scriptsize $\top$}}}}
\newcommand {\invtrans}		{\ensuremath{^{\mbox{\scriptsize $-\top$}}}}
\newcommand {\inverse}		{\ensuremath{^{\mbox{\scriptsize -1}}}}
\newcommand {\smallTH}		{\ensuremath{^{\mbox{\scriptsize th}}}}
\newcommand {\trace}        {\ensuremath{\mathrm{tr}}}
\newcommand {\RE}           {\ensuremath{\mathbb{R}}}
\newcommand {\strad}		{\bar{}\!\!\!\!\:d}
\newcommand {\bstrad}		{\,\,\,\bar{}\!\!\!\!\:d}
\newcommand {\xinit}		{\ensuremath{x^{\circ}}}
\newcommand {\optimal}		{\ensuremath{^{\mbox{\scriptsize \textsf {OPT}}}}}
\newcommand {\liealge}		{\ensuremath{\mathfrak{g}}}
\newcommand {\liea}			{\ensuremath{\mathfrak{g}}}

\newcommand {\etal}         {\textit{et al.} }
\newcommand {\eg}           {\textit{e.g.,} }
\newcommand {\ie}           {\textit{i.e.,}	}
\newcommand {\etc}          {\textit{etc} }
\newcommand {\GTpp}         {{\em G2++}	\xspace}
\newcommand {\ito}          {It\^{o}			\xspace}
\newcommand {\R}            {\hbox{I \kern -.5em R}}

\algdef{SE}[DOWHILE]{Do}{doWhile}{\algorithmicdo}[1]{\algorithmicwhile\ #1}%

\newcommand{\cpp}{C\nolinebreak\hspace{-.05em}\raisebox{.4ex}{\tiny\bf +}\nolinebreak\hspace{-.10em}\raisebox{.4ex}{\tiny\bf +}}
\def\cpp{{C\nolinebreak[4]\hspace{-.05em}\raisebox{.4ex}{\tiny\bf ++}}}
\begin{abstract}

  We present an efficient variational integrator for simulating
  multibody systems. Variational integrators reformulate the equations
  of motion for multibody systems as discrete Euler-Lagrange (DEL)
  equation, transforming forward integration into a root-finding
  problem for the DEL equation.  Variational integrators have been
  shown to be more robust and accurate in preserving fundamental
  properties of systems, such as momentum and energy, than many
  frequently used numerical integrators. However, state-of-the-art
  algorithms suffer from $O(n^3)$ complexity, which is prohibitive for
  articulated multibody systems with a large number of degrees of freedom, $n$, in generalized coordinates. Our key contribution is to
  derive a quasi-Newton algorithm that solves the root-finding problem
  for the DEL equation in $O(n)$, which scales up well for complex
  multibody systems such as humanoid robots. Our key insight is that
  the evaluation of DEL equation can be cast into a \emph{discrete inverse
  dynamic} problem while the approximation of inverse Jacobian can be
  cast into a \emph{continuous forward dynamic} problem.  Inspired by
  Recursive Newton-Euler Algorithm (RNEA) and Articulated Body
  Algorithm (ABA), we formulate the DEL equation individually for each
  body rather than for the entire system, such that both inverse and
  forward dynamic problems can be solved efficiently in $O(n)$.  We
  demonstrate scalability and efficiency of the variational integrator
  through several case studies.

\keywords{variational integrator $\cdot$ discrete mechanics $\cdot$ multibody systems $\cdot$ dynamics $\cdot$ computer animation \& simulation}

\end{abstract}

%
\section{Introduction}
%

We address the problem of accurately and efficiently simulating the dynamics of complex multibody systems, often referred to as the forward dynamics problem. Existing state-of-the-art approaches use the Lagrangian formalism, expressing the difference between kinetic and potential energy (the Lagrangian) in generalized coordinates, and derive the Euler-Lagrange second-order differential equations from them via the principle of least action. The state of the system at any time $t$ is then obtained by integrating these differential equations from initial conditions.

However, the long-term conservation of conserved quantities like energy and momentum of the system remains a key open challenge. In particular, discrete-time simulations, even with advanced algorithms for solving differential equations, eventually produce alarming and physically implausible behaviors, even for simple dynamical systems like $N$-link pendulums, due to the accumulation of numerical errors.

To address this problem, Marsden and West \cite{marsden2001discrete} introduced the discrete Lagrangian, which approximates the integral of the Lagrangian over a small time interval. They then derived its variation via the principle of least action, creating the discrete Euler-Lagrange (DEL) equations. They also showed that variational integrators based on the DEL formulation were symplectic (energy-conserving) and crucially decoupled energy behavior from step size \cite{marsden2001discrete,west2004variational}. 

Unfortunately, despite their benefits for stability, variational integrators suffer from computational complexity. Variational integrators transform the integration of the equations of motion into a root-finding problem for the DEL equation. This introduces complexity in three places as most nonlinear root-finding algorithms require: (1) the evaluation of the DEL equation, (2) computation of their gradient (Jacobian), and (3) the inversion of the gradient. Although there exist efficient algorithms for evaluating the DEL equation, they \emph{do not} use generalized coordinates but instead treat each link as a free-body and apply constraint forces to enforce joints \cite{betsch2006discrete,leyendecker2008variational,leyendecker2010discrete}. This becomes especially complicated with branching multi-body systems and joint constraints. 

Recently Johnson and Murphey \cite{johnson2009scalable} proposed a scalable variational integrator that represents the DEL equation in generalized coordinates. By representing the multibody system as a tree structure in generalized coordinates, they showed that the DEL equation, as well as the gradient and Hessian of the Lagrangian, can be calculated recursively. However, the complexity of their algorithm is $O(n^2)$ for evaluating the DEL equation, and $O(n^3)$ for computing the Jacobian. When coupled with traditional root-finders, \eg Newton's method, that require the inverse of the Jacobian, this adds an approximately $O(n^3)$ complexity for matrix inversion.

In this paper, we introduce a new variational integrator for multibody dynamic systems. The primary contribution is an $O(n)$ algorithm which solves the root-finding problem for the DEL equation. Our key insight is that the evaluation of DEL can be cast into a discrete inverse dynamics problem \cite{luh1980line,featherstone2014rigid} while the root updating can be cast into a continuous forward dynamics problem. Both inverse and forward dynamics problems can be solved efficiently in $O(n)$ using a recursive Lie group formulation of the dynamics \cite{park1995lie,lee2008computational,kobilarov2011discrete,kobilarov2009lie}.

Inspired by Recursive Newton-Euler Algorithm (RNEA) and Articulated Body Algorithm (ABA), we formulate the DEL equation individually for each body rather than for the entire system. By taking advantage of the recursive relations between body links, it becomes possible to evaluate the DEL function using a discrete inverse dynamics algorithm in linear-time. The same recursive representation is applied to update the root using an impulse-based forward dynamics algorithm. Together with these two algorithms, we propose an $O(n)$ quasi-Newton method specialized for finding the root of DEL equation, resulting in a \emph{Linear-Time Variational Integrator}.

We compare our method with the state-of-the-art variational integrator in generalized coordinates \cite{johnson2009scalable}. The results show that, for the same computation method of root updating, the performance of our recursive evaluation of the DEL equation (linear-time DEL algorithm) is 15 times faster for a system with 10 degrees of freedom (DOFs) and 32 times faster for 100 DOFs. For the same evaluation method of the DEL equation (\ie linear-time DEL algorithm), our results show that the performance of our new quasi-Newton method is 3.8 times faster for a system with 10 DOFs, and 53 times faster for 100 DOFs. Further analysis shows that for higher DOF systems, the impulse-based Jacobian approximation becomes increasingly more effective compared to our linear-time DEL algorithm.

%
\section{Background}
%

Our work is built on the concepts of discrete mechanics and
variational integrators. In this section, we will
briefly describe the standard formulation of discrete mechanics
\cite{johnson2009scalable}, followed by a reformulation using the Lie group 
representation for the Special Euclidean group $\SEthree$ of rigid body motions \cite{kobilarov2011discrete}.

\subsection{Variational Integrators in Generalized Coordinates}

We begin with the definition of Lagrangian, $L(\mathbf{q},
\dot{\mathbf{q}}) \in \mathbb{R}$,  the difference between the total kinetic energy
and the total potential energy of a system characterized by
generalized coordinates $\mathbf{q} \in \mathbb{R}^n$ where $n$ denotes the degrees of freedom of the system. For continuous-time systems, the principle of least action states that the system will follow the 
trajectory that minimizes the \textit{action integral} $\int_{t_1}^{t_2} L(\mathbf{q}(t), \dot{\mathbf{q}}(t)) dt$.

However, when we simulate the mechanical system on a
    computer, the mechanical system takes \textit{discrete time
    steps} rather than following the continuous trajectory.
Loosely speaking, the idea of discrete mechanics is that
  the system will follow the \textit{discretized trajectory} that
  minimizes the approximated action integral defined on the discretized
  trajectory. If we discretize a continuous trajectory $\mathbf{q}(t)$ into a sequence of configurations
$\mathbf{q}^0, \mathbf{q}^1, \cdots, \mathbf{q}^N$, we can define a
discrete Lagrangian that approximates the integral of $L(\mathbf{q}(t),
\dot{\mathbf{q}}(t))$ over a short interval $\Delta t$:
\begin{align}
L_d(\mathbf{q}^k, \mathbf{q}^{k+1}) \approx \int_{k \Delta t}^{(k+1) \Delta t} L(\mathbf{q}(t), \dot{\mathbf{q}}(t)) dt.
\label{eq:discrete_lagrangian}
\end{align}

Using the discrete Lagrangian, we can define the \textit{action sum}
$\sum_{k=0}^{N-1} L_d(\mathbf{q}^k, \mathbf{q}^{k+1})$ as an approximation of 
the action integral.
Minimizing the action sum with respect to $\{\mathbf{q}^k\}$ ($k=1, 2, \cdots, N-1$), 
we arrive at the discrete Euler-Lagrange (DEL) equation:
\begin{align}
D_2 L_d(\mathbf{q}^{k-1}, \mathbf{q}^k)
+ D_1 L_d(\mathbf{q}^k, \mathbf{q}^{k+1}) = 0,
\label{eq:DEL}
\end{align}
where $D_i: \RE \rightarrow \RE^n$ denotes differential operator with respect to the $i$-th parameter
of the function, and the differentials of $L_d$ can be analytically computed \cite{johnson2009scalable}.
Note that the boundary configurations $q^0$ and $q^N$ are not
  varied.

Instead of numerically integrating the Euler-Lagrange equation to simulate the
trajectory, discrete mechanics solves a \textit{root-finding problem} to obtain the next
configuration. Specifically, given two previous configurations $\mathbf{q}^{k-1}$ and
$\mathbf{q}^k$, we solve the next configuration $\mathbf{q}^{k+1}$ by
finding the root of the following function:
\begin{align}
f(\mathbf{q}^{k+1}) = D_2 L_d(\mathbf{q}^{k-1}, \mathbf{q}^k) 
                    + D_1 L_d(\mathbf{q}^k, \mathbf{q}^{k+1}) = 0.
\label{eq:DEL_function}
\end{align}

The superior energy behavior of variational integrators compared to
the traditional integrators like Euler and Runge-Kutta methods have been
shown using a discrete version of Noether's theorem \cite{marsden2001discrete}. 
One geometric
interpretation of variational integrators is that the DEL equation plays the
role of constraints, enforcing the discrete system to evolve on the constraint
manifold such that $f(\mathbf{q}^{k+1}) = 0$, \ie satisfying the least action principle
on the approximated action. In that sense, the process of root-finding can
be seen as a feedback controller to find the physically correct configuration for
the next time step, with the DEL equation being used by the feedback law to
indicate how far away the given configuration is from the manifold.  Traditional
integrators do not have such indicators, only account for the rate of change
based on the current state, which leads to the numerical error accumulation.

This nonlinear, high-dimensional, continuous root-finding problem can be solved
efficiently by Newton's method, provided that the partial derivatives of
$f$, $J_f(\mathbf{q})$ (\ie the Jacobian matrix), can be evaluated:

\begin{algorithm}[H]
	\caption{Newton's Method for Solving DEL Equation}\label{alg:root_finding}
	\begin{algorithmic}[1]
		\State \textbf{Initial Guess $\mathbf{q}_{0}$}
		\Do
		\State Evaluate $f(\mathbf{q}^{k+1})$
                \Comment{$O(n^2)$ time}
                \State \textbf{if}  $\|f(\mathbf{q}^{k+1}) < \epsilon\|$
                \;\;\; \Return $\mathbf{q}^{k+1}$
		\State Update $\mathbf{q}^{k+1} \leftarrow \mathbf{q}^{k+1} -\left[J_f(\mathbf{q}^{k+1})\right]^{-1} f(\mathbf{q}^{k+1})$ \Comment{$O(n^3)$ time}
		\doWhile{num\_iteration $<$ max\_iteration} 
	\end{algorithmic}
\end{algorithm}

To avoid the computation of the Jacobian and its inversion, various
quasi-Newton methods can be applied to approximate
$\left[J_f(\mathbf{q}^{k+1})\right]^{-1}$. In Section \ref{sec:linear-time jacobian approximation}, we introduce a
linear-time algorithm to approximate the product of
  $\left[J_f(\mathbf{q}^{k+1})\right]^{-1}$ and $
  f(\mathbf{q}^{k+1})$ for finding the root of DEL equation.

\subsection{Variational Integrators in $\SEthree$}

The linear-time root-finding algorithm we will introduce in the next
section leverages the idea of reformulating DEL equation for each rigid body rather than for the entire system.
We begin with the expression of the DEL equation in $\SEthree$ for a single rigid body.

The configuration of the rigid body can be represented by matrices of the form:
\begin{align}
T = \begin{bmatrix}
R & p \\ 0 & 1
\end{bmatrix} \in \SEthree,
\end{align}
where $R \in \SOthree$ is a $3 \times 3$ rotation matrix, and $p \in \mathbb{R}^3$ is a position vector. 
The spatial velocity of the rigid body $V = (w, v) \in \sethree$ or twist can be represented in six-dimensional vector or $4 \times 4$ matrix form:
\begin{align}
V = \begin{pmatrix} w \\ v \end{pmatrix},
~~
[V] = 
\begin{bmatrix}
\hat{w} & v \\
0 & 0
\end{bmatrix},
\end{align}
where $w \in \sothree$ and $v \in \RE^3$ denote the angular velocity and linear velocity, respectively, and $\hat{w}$ is the $3\times3$ skew symmetric matrix for $w$ such that $\hat{w}^T = -\hat{w}$. In this paper, we use
brackets $[\cdot]$ to denote matrix representations.

The Lagrangian of a rigid body can be compactly expressed using the Lie group
representation (\cite{park1995lie,murray1994mathematical}) in the space
of $\SEthree$:
\begin{align}
L(T, V) = \frac{1}{2} V^T G V - P(T),
\label{eq:continuous_lagrangian_of_single_body}
\end{align}
where $P: \SEthree \rightarrow \mathbb{R}$ is the potential energy. $G$ is the spatial inertia matrix that has the following structure:
\begin{align}
G = \begin{bmatrix}
\I{} & 0 \\
0 & m I
\end{bmatrix} \in \mathbb{R}^{6 \times 6},
\label{eq:spatial_inertia_matrix}
\end{align}
where $\I{}$ is the inertia matrix, $m$ is the mass, and $I$ is 
$3 \times 3$ identity matrix when the center of mass is at the origin of the 
body frame.

Analogous to Equation \eqref{eq:discrete_lagrangian}, the discrete Lagrangian for a single rigid body can be expressed as
\begin{align}
L_d(T^k, T^{k+1}) \approx \int_{k \Delta t}^{(k+1) \Delta t} L(T, V) dt.
\label{eq:discrete_lagrangian_of_single_body}
\end{align}

In this paper, we use the trapezoidal quadrature approximation for the discrete Lagrangian of the single body system as
\begin{align}
L_d(T^k, T^{k+1}) \defeq \frac{\Delta t}{2} L(T^k, V^k) + \frac{\Delta t}{2} L(T^{k+1}, V^k),
\label{eq:discrete_lagrangian_trapezoidal}
\end{align}
where the \textit{average velocity} $V^k$ can be defined as
\begin{align}
V^k = \frac{1}{\Delta t} \log ( \Delta T^k )
\label{eq:average_velocity_of_single_body},
\end{align}
with the \textit{log map} $\log: \SEthree \rightarrow \sethree$, the inverse of the \textit{exponential map} $\exp: \sethree \rightarrow \SEthree$
\cite{kobilarov2011discrete,murray1994mathematical}, and $\Delta T^k = \left.T^k\right.^{-1} T^{k+1}$, the displacement
of the rigid body's configuration during the discrete times of $t_k$ and $t_{k+1}$.

To derive the DEL equation for a single rigid body in $\SEthree$, we need
to take the variational calculus on $V^k$ with respect to $T^k$ and $T^{k+1}$. This requires
the derivative of log map defined as
\begin{align}
\label{eq:inverse_iexp}
\left( \frac{\partial}{\partial T} \log(T) \right) [W] &= \dlog_{V} \left( [W] \exp(-[V]) \right),
\end{align}
where $V = \log(T)$, and $W \in \sethree$ is an arbitrary twist, and $\dlog_V: \sethree \rightarrow \sethree$ is the inverse of the right trivialized tangent $\dexp_V: \sethree \rightarrow \sethree$ as an linear operator \cite{kobilarov2011discrete,bou2009hamilton}:
\begin{align}
\dlog_V (W) &= \sum_{j = 0}^{\infty} \frac{B_j}{j!} \ad{V}^j (W).
\end{align}

The Lie bracket operator $\ad{V}{}: \sethree \rightarrow \sethree$ is defined as $\ad{V}(W) = [V][W] - [W][V]$. $\dlog_V$ can be alternatively represented in matrix form as
\begin{align}
[\dlog_V] = \sum_{j = 0}^{\infty} \frac{B_j}{j!} [\ad{V}]^j,
~~~~
[\ad{V}] = \begin{bmatrix}
\hat{w} & 0 \\
\hat{v} & \hat{w}
\end{bmatrix},
\end{align}
where $B_j$ are the Bernoulli numbers ($B_0 = 1, B_1 = -1/2, B_2 = 1/6, B_3 = 0, \dots$) \cite{hairer2006geometric}.

Using Equation \eqref{eq:average_velocity_of_single_body} and \eqref{eq:inverse_iexp}, we can now express the variation of ${V^k}$ as
\begin{align}
  \delta V^{k} = \frac{1}{\Delta t} \dlog_{\Delta t V^k} \left(
  -\left.T^k\right.^{-1} \delta T^k + \Ad{\exp(\Delta t [V^k])} \left( \left.T^{k+1}\right.^{-1}
\delta T^{k+1} \right) \right),
\label{eq:deltaV}
\end{align}
where $\delta T^k$ and $\delta T^{k+1}$ are variations, and $\AdT{}: \sethree \rightarrow \sethree$ is the adjoint action of
$T \in \SEthree$ on $V \in \sethree$ defined as $\AdT{}V = T [V]
T^{-1}$. The adjoint action can be regarded as an linear operator in the $6 \times 6$ matrix form of:
\begin{align}
[\AdT{}] = \begin{bmatrix}
R & 0 \\ \hat{p}R & R
\end{bmatrix}.
\end{align}

By the least action principle with Equation \eqref{eq:discrete_lagrangian_trapezoidal},
\eqref{eq:average_velocity_of_single_body}, and \eqref{eq:deltaV}, we can
derive the DEL equation for a single rigid body in $\SEthree$, which
is the well known \textit{discrete reduced Euler-Poincar{\'e}} equations \cite{kobilarov2011discrete,fan2015structured}:
\begin{subequations}
\begin{align}
D_2 L_d(T^{k-1}, T^k) + D_1 L_d(T^k, T^{k+1}) = 0 ~~ \in \RE^6,
\label{eq:discrete_reduced_euler_poincare_a}
\end{align}
\text{where}
\begin{align}
D_2 L_d(T^{k-1}, T^k) &= - [\Ad{\exp(\Delta t [V^{k-1}])}]^T \left[ \dlog_{\Delta t V^{k-1}} \right]^T G V^{k-1} + \frac{\Delta t}{2} \left. T^k \right.^* P(T^k)
\label{eq:discrete_reduced_euler_poincare_b}\\
D_1 L_d(T^k, T^{k+1}) &= \left[ \dlog_{\Delta t V^k} \right]^T G V^k + \frac{\Delta t}{2} \left. T^k \right.^* P(T^k).
\label{eq:discrete_reduced_euler_poincare_c}
\end{align}
\end{subequations}

By Lagrange-d'Alembert principle, Equation 
\eqref{eq:discrete_reduced_euler_poincare_a} can be straightforwardly extended
to a forced system \cite{kobilarov2011discrete}:
\begin{align}
D_2 L_d(T^{k-1}, T^k) + D_1 L_d(T^k, T^{k+1}) + F^{k} = 0,
\label{eq:forced_del}
\end{align}
where $F^{k} \in \dsethree$ is the integral of the virtual work performed by the force over the time interval $\Delta t$.

%
\section{Linear-Time Variational Integrator}
%

\begin{figure}[t]
	\begin{subfigure}{.5\textwidth}
		\centering
		\includegraphics[width=50mm]{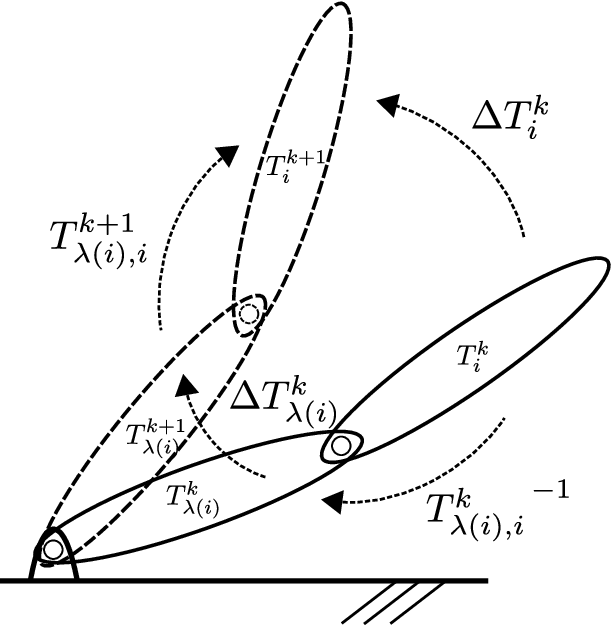}
		\caption{Displacement of body $i$'s configuration}
		\label{fig:recursive_displacement}
	\end{subfigure}
	\begin{subfigure}{.5\textwidth}
		\includegraphics[width=65mm]{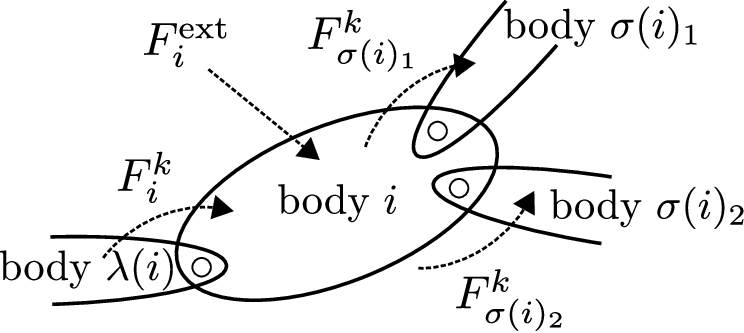}
		\caption{Impulses acting on body $i$}
		\label{fig:recursive_average_newton_euler}
	\end{subfigure}
	\caption{Recurrence relationships of configuration displacement and impulses}
	\label{fig:recursive}
\end{figure}

We introduce a new linear-time variational integrator which, at each
time instance $t_k$, solves for the root of Equation
\eqref{eq:DEL_function}. Our variational integrator consists of two
linear-time algorithms for evaluating the DEL equation and updating the
root, which, as shown in Algorithm \ref{alg:root_finding}, determine the
time complexity of the root-finding algorithm. We first derive the DEL
equation for multibody systems in a recursive manner, resulting a
linear-time procedure to evaluate the function $f(\mathbf{q})$. Next,
we introduce an impulse-based dynamics algorithm, which is also
linear-time, to estimate the next configuration. Replacing Line 3 and Line 5
in Algorithm \ref{alg:root_finding} with these two
algorithms, we present a new linear-time quasi-Newton root-finding
method for finding the root of DEL equation.

\subsection{Linear-Time Evaluation of the DEL Equation}

If we view the function $f(\mathbf{q}) = \mathbf{0}$ as a dynamic
constraint that enforces the equation of motion, any nonzero value of
$f(\mathbf{q})$ indicates the residual impulse that violates the
equation of motion. As such, evaluating $f(\mathbf{q})$ can be
considered a discrete inverse dynamics problem which solves the
residual impulse of the system given $\mathbf{q}^{k-1}$,
$\mathbf{q}^{k}$, and $\mathbf{q}^{k+1}$. We derive a recursive DEL
equation using similar formulation as recursive Newton-Euler algorithm (RNEA)
\cite{luh1980line,featherstone2014rigid}, 
which solves the inverse
dynamics for continuous systems in linear time 
with respect to the degrees of freedom of the system.

Assuming that the multibody system can be represented as a
tree-structure where each body has at most one parent and an arbitrary
number of children, connected by joints, our goal is to expand Equation
\eqref{eq:forced_del} to account for the dynamics of entire
tree-structure.

We begin with the recursive definition for a rigid body's configuration and the
displacement of the configuration. Let us denote $\{0\}$ as an inertial
frame which is stationary in the space, $\{i\}$ as body frame of
$i$-th body in the tree structured system, and $\{\lambda(i)\}$ as a
body frame of the parent of the $i$-th body. The configuration of a
body in the system can be represented as
\begin{align}
	T_{i}^{k} &= T_{\lambda(i)}^{k} T_{\lambda(i),i}^{k},
	\label{eq:multibody_transformation}
\end{align}
where $T_{i}^{k}$ and $T_{\lambda(i)}^k$ denote the transformations 
from the inertial frame to $\{i\}$ and $\{\lambda(i)\}$, respectively, while 
$T_{\lambda(i), i}^{k}$ denotes the relative transformation from $\{\lambda(i)\}$ to $\{i\}$
represented as a function of the $i$-th joint configuration $\mathbf{q}_i^{k+1}$.
From Equation \eqref{eq:multibody_transformation}, the configuration displacement
of a rigid body can be written as
\begin{align}
\Delta T_{i}^{k} = \left.T_{\lambda(i),i}^{k}\right.^{-1} 
\Delta T_{\lambda(i)}^{k} T_{\lambda(i),i}^{k+1}.
\label{eq:recursive_configuration_displacement}
\end{align}

Fig. \ref{fig:recursive} (a) gives a geometric interpretation of the
recurrence relationship of the configuration displacements between 
$\Delta T_{i}^{k}$ and $\Delta T_{\lambda(i)}^{k}$.

Plugging Equation \eqref{eq:recursive_configuration_displacement} into Equation \eqref{eq:average_velocity_of_single_body}, we can obtain the average velocity of $i$-th
rigid body as $V_i^k = \frac{1}{\Delta t}\log(\Delta T_{i}^{k})$. 
Unlike the continuous velocity of $i$-th body $V_i = S_i \dot{\mathbf{q_i}}$ where $S_i$ is the
joint Jacobian \cite{murray1994mathematical},
the equation for the average velocity is implicit with respect to $\mathbf{q}^{k+1}$ due to
the log map.
The use of log map, with $\dlog_V$, is the key reason that makes the DEL equation implicit with respect to
$\mathbf{q}^{k+1}$.

For a rigid body in a multibody system, the impulse term $F^k$ in
Equation \eqref{eq:forced_del} includes the impulse transmitted from the
parent link $F_i^k$, impulses transmitting to the child links
$F_c^k$, and other external impulses $F_i^{\text{ext},k}$ applied by
the environment as (Fig. \ref{fig:recursive} (b)):
\begin{align}
F^k = F_i^k 
  - \sum_{c \in \sigma(i)} 
    \dAd{ \left. T_{i, c}^k \right.^{-1} }{F_{c}^{k}} 
  + F_i^{\text{ext},k}.
\label{eq:sum_of_forces}
\end{align}

Note that $F_i^k$ is expressed in the $i$-body coordinates so the 
coordinate frame transformation is required for $F_c^k$ as
$\left[ \Ad{ \left. T_{i, c}^k \right.^{-1} } \right]^T {F_{c}^{k}}$.

Plugging these forces into Equation
\eqref{eq:forced_del} and using the definitions in Equation
\eqref{eq:discrete_reduced_euler_poincare_b} and
\eqref{eq:discrete_reduced_euler_poincare_c}, we express the equations of motion for the i-th body as
\begin{subequations}
\begin{align}
F_{i}^k &= \mu_{i}^{k} - \left[ \Ad{\exp(\Delta t [V_{i}^{k-1}])} \right]^T {\mu_{i}^{k-1}} + \sum_{c \in \sigma(i)} \left[ \Ad{ \left. T_{i, c}^k \right.^{-1} } \right]^T {F_{c}^{k}} - F_i^{\text{ext},k} \\
\mu_i^k &= \left[ \dlog_{\Delta t V_i^k} \right]^T G_i V_i^k,
\label{eq:reduced_euler_poincare_for_multibody_systems}
\end{align}
\end{subequations}
where $\mu_i^k$ is the
discrete momentum of body $i$ and $\sigma(i)$ denotes the set of child bodies to
body $i$. The required generalized impulse of joint $i$ to achieve the motion 
$q^{k+1}$ is simply the projection of $F_i^k$ onto the joint Jacobian as 
$S_i^T F_i^k$ where $S_i \in \RE^{6 \times n_i}$ is the $i$-th joint Jacobian
\cite{murray1994mathematical}. The residual impulse then can be obtained by 
subtracting the joint impulses, $Q_i^k$, such as joint actuation or joint friction, from the required impulse:
\begin{align}
f_i = S_i^T F_i^k - Q_i^k ~~~~ \in \RE^{n_i}.
\label{eq:generalized_force}
\end{align}

Algorithm \ref{alg:drnea} summarizes the recursive procedure, which we
call discrete recursive Newton-Euler algorithm
(DRNEA). DRNEA consists a forward pass from the root of the
tree structure to the leaf nodes and a backward pass in the reverse
order. The forward pass computes the velocity of each body
while the backward pass computes force transmitted between
joints. By exploiting the recursive relationship between a parent
body and its child bodies, the computation for each pass is $O(n)$, where
$n$ is the number of rigid body links in the system assuming 
the degree of freedom of each joint is one.

\begin{algorithm}[H]
	\caption{Discrete recursive Newton-Euler algorithm (DRNEA)}\label{alg:drnea}
	\begin{algorithmic}[1]
		\For {$i = 1 \to n$}
		\State {$T_{\lambda(i),i}^{k+1} = \text{function of } q_i^{k+1}$}
		\State {$\Delta T_{i}^{k} = \left.T_{\lambda(i),i}^{k}\right.^{-1} \Delta T_{\lambda(i)}^{k} T_{\lambda(i),i}^{k+1}$}
		\State {$V_i^k = \frac{1}{\Delta t} \log \left( \Delta T_{i}^{k} \right)$}
		\EndFor
		\For {$i = n \to 1$}
		\State {$\mu_{i}^{k} = \left[ \dlog_{\Delta t V_{i}^{k}} \right]^T G_i V_{i}^{k}$}
		\State {$F_{i}^k = \mu_{i}^{k} - \left[ \Ad{\exp(\Delta t [V_{i}^{k-1}] )} \right]^T {\mu_{i}^{k-1}} - F_i^{\text{ext},k} + \sum_{c \in \sigma(i)} \left[ \Ad{ \left. T_{i, c}^k \right.^{-1} } \right]^T {F_{c}^{k}}$}
		\State {$f_{i} = S_{i}^{T} F_i^k - Q_{i}^{k}$} 
		\EndFor
	\end{algorithmic}
\end{algorithm}

For clarity, the mathematical symbols used in DRNEA are listed below.

\begin{itemize}[label={$\bullet$}]
	\item $i$: index of the $i$-th body.
	\item $\lambda(i)$: index of the parent body of the $i$-th body.
	\item $\sigma(i)$: set of indices of the child bodies of the $i$-th body.
	\item $q_i^k \in \RE^{n_i}$: generalized coordinates of the
          $i$-th joint which connects the $i$-th body with its parent
          body where $n_i$ denotes the dimension of the coordinates.
	\item $Q_i \in \RE^{n_i}$: generalized force exerted by the $i$-th joint.
	\item $T_{\lambda(i), i} \in \SEthree$: relative transformation matrix from the $\{\lambda(i)\}$ to $\{i\}$.
	\item $V_i^k \in \sethree$: the spatial average velocity of the $i$-th body, expressed in \{i\} at time step $k$
	\item $S_i^k \in \RE^{6 \times {n_i}}$: Jacobian of $T_{\lambda(i), i}$ expressed in $\{i\}$.
	\item $G_i \in \RE^{6 \times 6}$: the spatial inertia of the $i$-th body, expressed in $\{i\}$.
	\item $F_i^k \in \dsethree$: the spatial impulse transmitted to the $i$-th body from its parent through the connecting joint, expressed in $\{i\}$.
	\item $F_i^{\text{ext},k} \in \dsethree$: the spatial impulse acting on the $i$-th body, expressed in $\{i\}$.
\end{itemize}

\subsection{Linear-Time Root Updating}
\label{sec:linear-time jacobian approximation}

Besides function evaluation, Newton-like methods also require the
update of Jacobian to estimate the root, which is usually the computation bottleneck in
each iteration. Here we describe a recursive impulse-based method to
efficiently update the root in linear-time.

Let us denote the current iteration in Newton's method as $l$ and the
current estimate of the configuration at next time step as
$\mathbf{q}_{(l)}^{k+1}$. Evaluating the forced DEL equation
\eqref{eq:forced_del} gives the residual impulse,
$f(\mathbf{q}_{(l)}^{k+1}) = \mathbf{e}_{(l)}$, in the system. If the
magnitude of $\mathbf{e}_{(l)}$ is zero or less than the tolerance,
$\mathbf{q}_{(l)}^{k+1}$ is the next configuration that satisfies the
forced DEL equation. Otherwise, $\mathbf{e}_{(l)}$ can be regarded as the
residual impulse needed to result in $\mathbf{q}_{(l)}^{k+1}$ at the
next time step. If we apply the negative residual force,
$-\mathbf{e_{(l)}}/\Delta t$, 
to the system, we should arrive at a
configuration closer to the root of
$f(\mathbf{q}^{k+1})$. Applying such a force to
the system can be done by continuous forward dynamics in linear-time \cite{featherstone2014rigid}.

Given the approximation of $\dot{\mathbf{q}}^k$ as
$\frac{1}{\Delta t}\left(\mathbf{q}^{k} - \mathbf{q}^{k-1}\right)$, the continuous
forward dynamics equation can be used to evaluate the generalized
acceleration:
\begin{align}
\ddot{\mathbf{q}}^k = M^{-1}(\mathbf{q}^k) \left( -C(\mathbf{q}^k, \dot{\mathbf{q}}^k) \dot{\mathbf{q}}^k + Q \right),
\label{eq:continuous_eom}
\end{align}
where $M(\mathbf{q}^k)$ is the mass matrix and $C(\mathbf{q}^k,
\dot{\mathbf{q}}^k)$ is the Coriolis force in generalized
coordinates. $Q$ indicates the sum of other external and internal
forces applied to the system in generalized coordinates.

Using the 2nd order central difference to approximate $\mathbf{q}^{k+1} =
\Delta t^2 \ddot{\mathbf{q}}^k + 2 \mathbf{q}^{k} - \mathbf{q}^{k-1}$,
we can apply the negative residual force to improve the estimate of root:
\begin{align}
\mathbf{q}_{(l+1)}^{k+1} = \Delta t^2 M^{-1}(\mathbf{q}^k) \left( -C(\mathbf{q}^k,\dot{\mathbf{q}}^k) \dot{\mathbf{q}}^k + Q 
             - \sum_{m=0}^{l} \frac{\mathbf{e}_{(m)}}{\Delta t} \right) + 2 \mathbf{q}^{k}
             - \mathbf{q}^{k-1}.
\label{eq:update_q_using_forward_dynamics}
\end{align}

Consolidating the quantities on the RHS of Equation
\eqref{eq:update_q_using_forward_dynamics} gives the update rule for $\mathbf{q}^{k+1}$:
\begin{align}
\mathbf{q}_{(l+1)}^{k+1} &= \mathbf{q}_{(l)}^{k+1} -
                           \Delta t M^{-1}(\mathbf{q}_{(l)}^{k})
                           \mathbf{e}_{(l)}, 
\label{eq:new_update_rule}
\end{align}
where $\Delta t M^{-1}(\mathbf{q}_{(l)}^{k}) \mathbf{e}_{(l)}$ can be
evaluated in $O(n)$ using recursive impulse-based dynamics (ABI
algorithm: articulated body inertia algorithm)
introduced by Featherstone \cite{featherstone2014rigid}. Specifically,
ABI is a forward dynamics algorithm which computes Equation
\eqref{eq:continuous_eom}. If we set
$\dot{\mathbf{q}} \equiv \mathbf{0}$ (to eliminate the Coriolis force)
and $Q \equiv \Delta t \mathbf{e}_{(l)}$, ABI will return exactly
$\Delta t M^{-1}(\mathbf{q}_{(l)}^{k}) \mathbf{e}_{(l)}$.

Comparing to the Newton's method in Algorithm \ref{alg:root_finding}, the inverse of Jacobian
matrix is approximated by the inverse mass matrix multiplied by $\Delta
t$. We name this algorithm RIQN (Recursive Impulse-based Quasi-Newton
method) and summarize it in Algorithm \ref{alg:new_quasi_newton_method}.

\begin{algorithm}[H]
	\caption{Recursive Impulse-based Quasi-Newton
method (RIQN)}\label{alg:quasi_newton}
	\begin{algorithmic}[1]
		\State \textbf{Initial Guess $\mathbf{q}_{0}^{k+1}$}
		\Do
		\State Use DRNEA to evaluate $\mathbf{e} \leftarrow
                f(\mathbf{q}^{k+1})$ \Comment{$O(n)$ time}
                \State \textbf{if} $\|\mathbf{e} < \epsilon\|$ \;\;\; \Return $\mathbf{q}^{k+1}$
		\State Use ABI to compute $\Delta t
                M^{-1}(\mathbf{q}^{k}) \mathbf{e}$ \Comment{$O(n)$ time}
		\State Update $\mathbf{q}^{k+1} \leftarrow \mathbf{q}^{k+1} -
                           \Delta t M^{-1}(\mathbf{q}^{k})
                           \mathbf{e}$
		\doWhile{ num\_iteration $<$ max\_iteration} 
	\end{algorithmic}
	\label{alg:new_quasi_newton_method}
\end{algorithm}

\subsection{Initial Guess}
\label{sec:initial_guess}
Similar to other Newton-like methods, our algorithm requires the
initial guess to be sufficiently close to the solution. We propose
three different ways to produce an initial guess for RIQN. 
\begin{itemize}[label={$\bullet$}]
\item \textrm{IG1}: Directly use the current configuration as the
  initial guess of the next configuration: $\mathbf{q}_{(0)}^{k+1} = \mathbf{q}^k$.
\item \textrm{IG2}: Apply explicit Euler integration, $\mathbf{q}_{(0)}^{k+1} =
  \mathbf{q}^{k} + \Delta t ~ \dot{\mathbf{q}}^{k}$, where
  $\dot{\mathbf{q}}^{k}$ is approximated by 
  $\frac{1}{\Delta t}\left(\mathbf{q}^{k} - \mathbf{q}^{k-1}\right)$.
\item \textrm{IG3}: Compute the acceleration via the equations
  of motion, $\ddot{\mathbf{q}}^{k} = M^{-1} \left(-C + Q)
  \right)$, and apply
  semi-implicit Euler integration to integrate velocity, $\dot{\mathbf{q}}^{k+1} 
		= \dot{\mathbf{q}}^{k} + \Delta t ~ \ddot{\mathbf{q}}^{k}$, followed by
                position, $\mathbf{q}_{(0)}^{k+1} = \mathbf{q}^{k} + \Delta t ~ \dot{\mathbf{q}}^{k+1}$.
\end{itemize}

%
\section{Experimental Results}
%

In this section, we describe the implementation of the proposed algorithms,
RIQN and DRNEA, and verify the algorithms in terms of efficiency and scalability
by comparing them to the state-of-are algorithms through case studies. We 
used fixed time step of 1 millisecond for all the experiments.

\subsection{Implementation}

The algorithms introduced by this paper and several state-of-art algorithms were 
implemented on top of DART \cite{dart,GVU-TR}, which is an $\cpp$ open source 
dynamics library for multibody systems. All of the simulations were performed on
a Intel Core i7-4970K @ 4.00 GHz desktop computer.

All the source code of the implementations is available at the GitHub repository.\footnote{\url{https://github.com/jslee02/wafr2016}}

\subsection{Energy Conservation}

\begin{figure}[t]
	\begin{subfigure}{.5\textwidth}
		\centering
		\includegraphics[width=40mm]{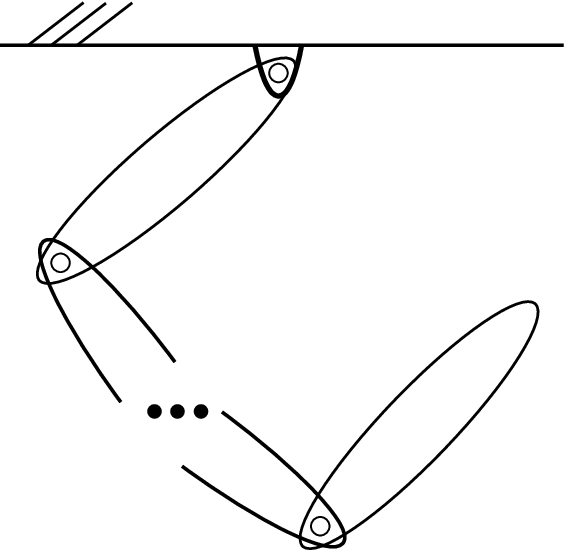}
	\end{subfigure}
	\begin{subfigure}{.5\textwidth}
		\centering
		\includegraphics[width=70mm]{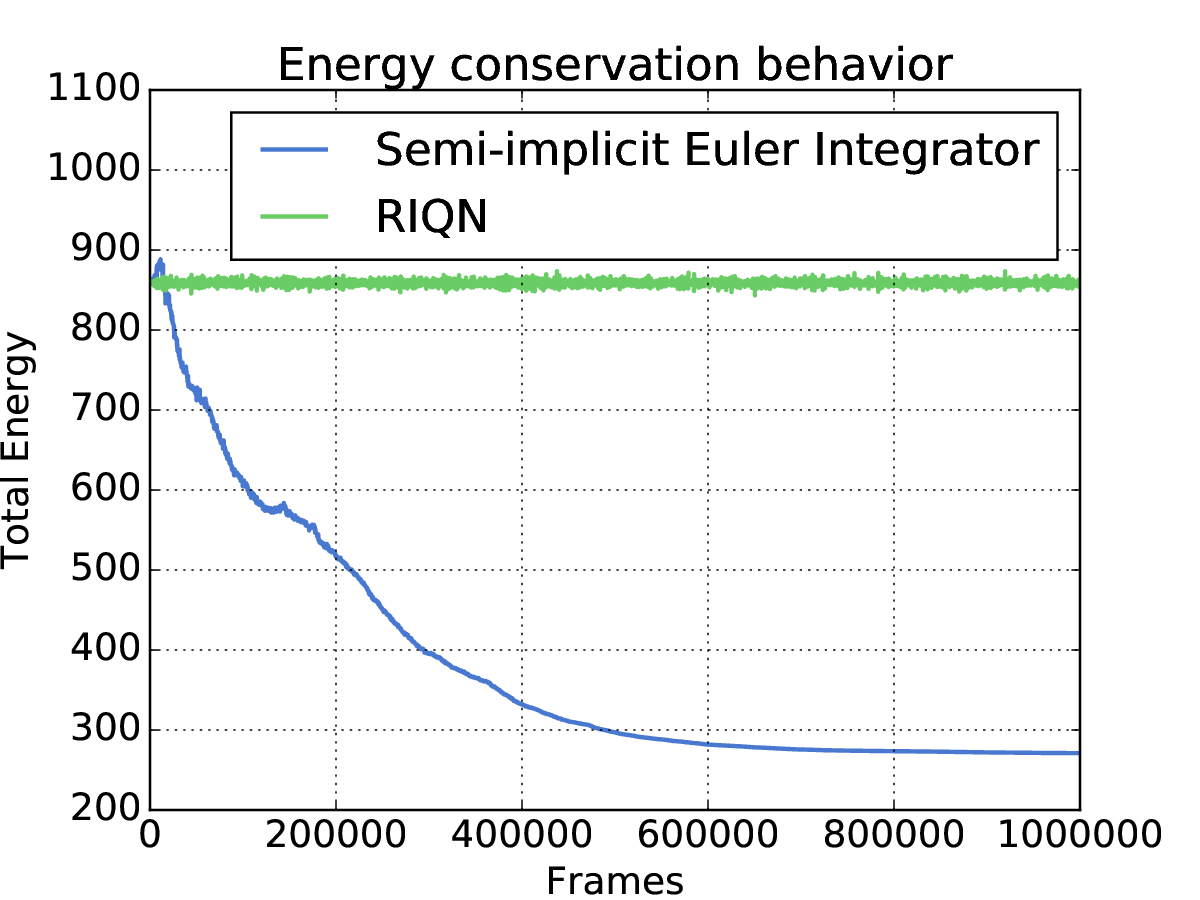}
	\end{subfigure}
	\caption{(a) Serial chain of $N$-bodies connected by revolute joints,
		(b) Energy conservation behavior over simulation frames}
	\label{fig:2}
\end{figure}

We first show that our linear-time variational integrator inherits the energy conservation property, which is one of the important features of variational integrators. We simulate a serial chain that consists of $N$-bodies connected by revolute joints (Fig. \ref{fig:2} (a)) with RIQN (variational integrator) and semi-implicit Euler method, which is an easy-to-implement standard method. In this experiment, we use a 10-body serial chain with no joint actuation nor external forces except for the gravity. The total energy (kinetic energy + potential energy) of this passive system should remain constant.

Fig. \ref{fig:2} (b) shows the energy evolution of the serial
chain over simulation frames for both integration methods. RIQN does not artificially dissipate the energy while the Euler method does.

\subsection{Performance Comparisons}
\label{sec:performance}

\begin{figure}[t]
	\begin{subfigure}{.5\textwidth}
		\centering
		\includegraphics[width=65mm]{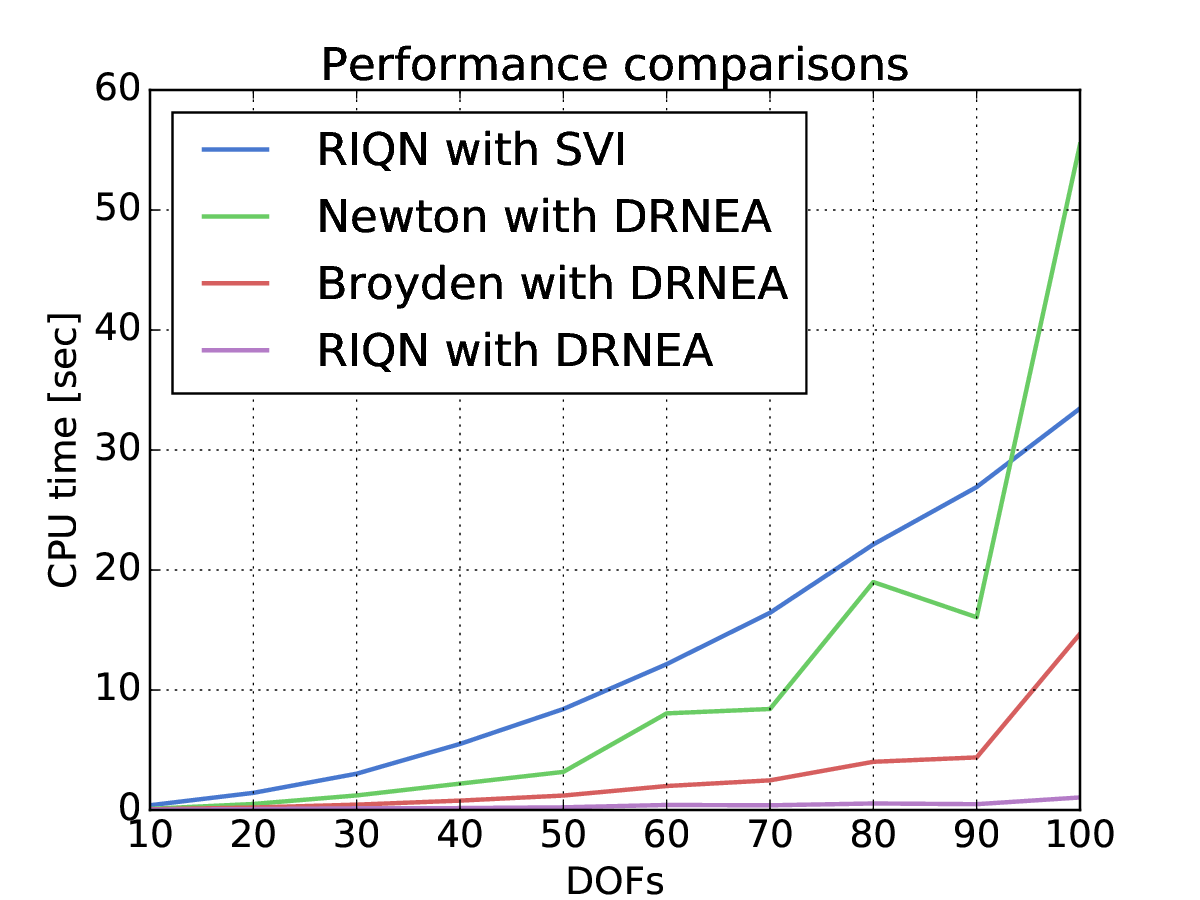}
		\caption{Performance comparisons}
		\label{fig:performance}
	\end{subfigure}
	\begin{subfigure}{.5\textwidth}
		\centering
		\includegraphics[width=65mm]{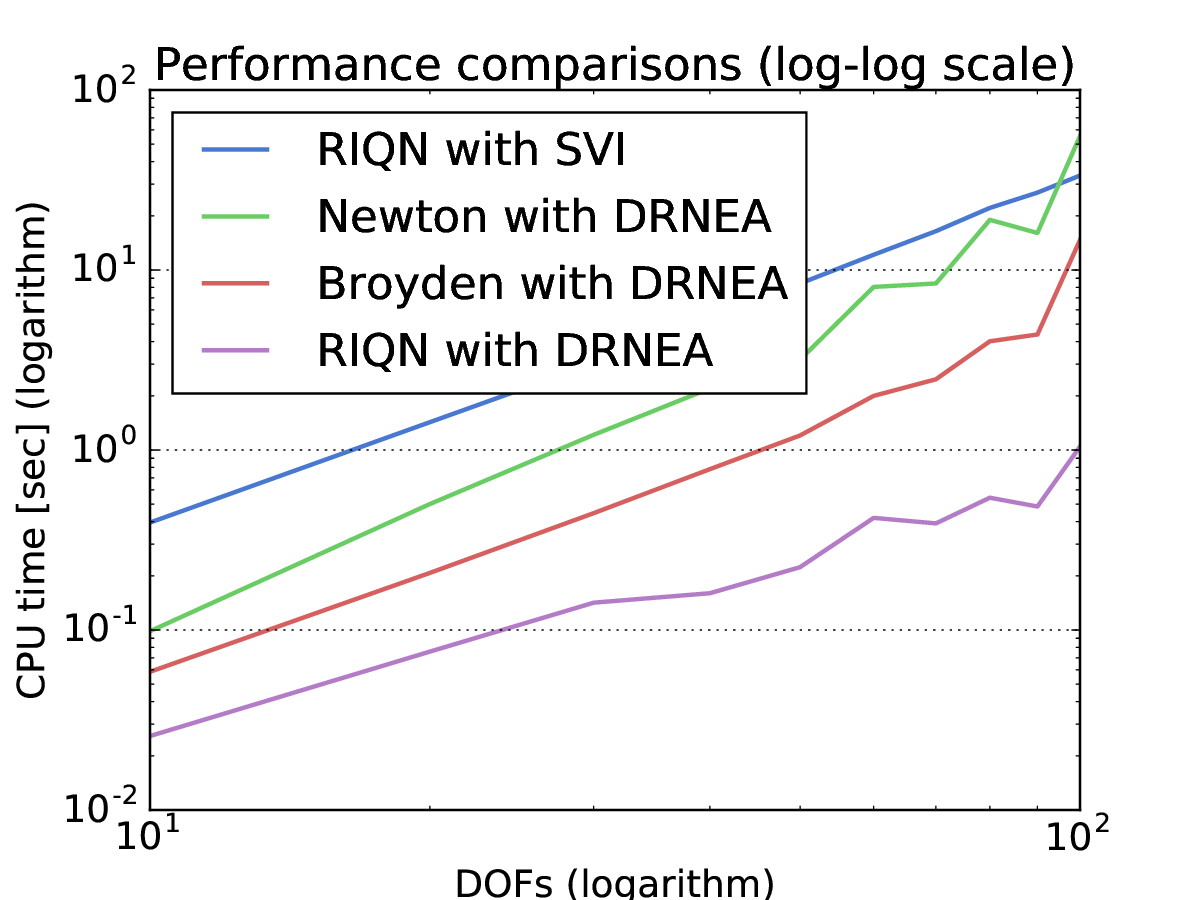}
		\caption{Performance comparisons (logarithm)}
		\label{fig:performance_log}
	\end{subfigure}
	\caption{Absolute computation time versus DOFs for the various 
		root-finding methods.}
	\label{fig:3}
\end{figure}

The major factors that affect on the computational time of 
variational integrator are (1) evaluation of DEL equation and (2) the evaluation of Jacobian inverse. We consider various of the root-finding algorithm that are combination of methods for (1) and (2).

For (1), we compare our DRNEA to the scalable variational integrator (SVI) \cite{johnson2009scalable}. For (2), we compare the proposed RIQN to Newton's method and Broyden method (quasi-Newton method) \cite{broyden1965class}. 

Newton's method requires the (exact) Jacobian of the DEL equation. When combining with DRNEA, for a fair comparison we also derive a recursive algorithm to evaluate the derivatives of the DEL equation with respect to $\mathbf{q}^{k+1}$. Please see the Appendix for the algorithm.

For all the root-finding methods, we measure computation time of serial chain forward 
dynamics simulations for 10k frames. To reveal the scalability of the methods, we vary the number of 
bodies of the serial chain (Fig. \ref{fig:3}). RIQN method with DRNEA shows the best performance. We also noticed that, for the
same method for (2), DRNEA shows better performance than SVI. Further analyses show that the impulse-based Jacobian approximation contributes more than our linear-time DEL algorithm for the higher DOFs systems.

\subsection{Convergence}

\begin{figure}[t]
	\begin{subfigure}{.5\textwidth}
		\centering
		\includegraphics[width=65mm]{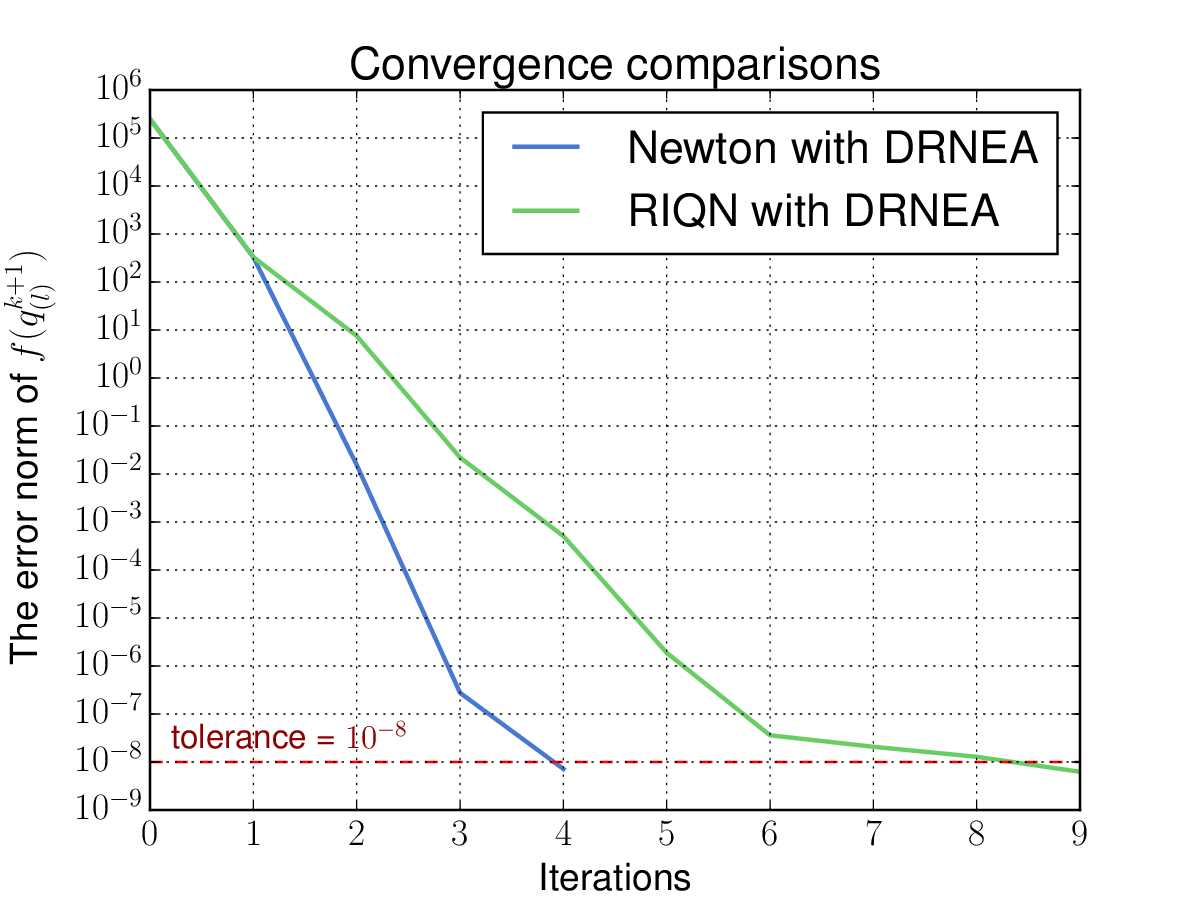}
		\caption{}
		\label{fig:convergence}
	\end{subfigure}
	\begin{subfigure}{.5\textwidth}
		\centering
		\includegraphics[width=65mm]{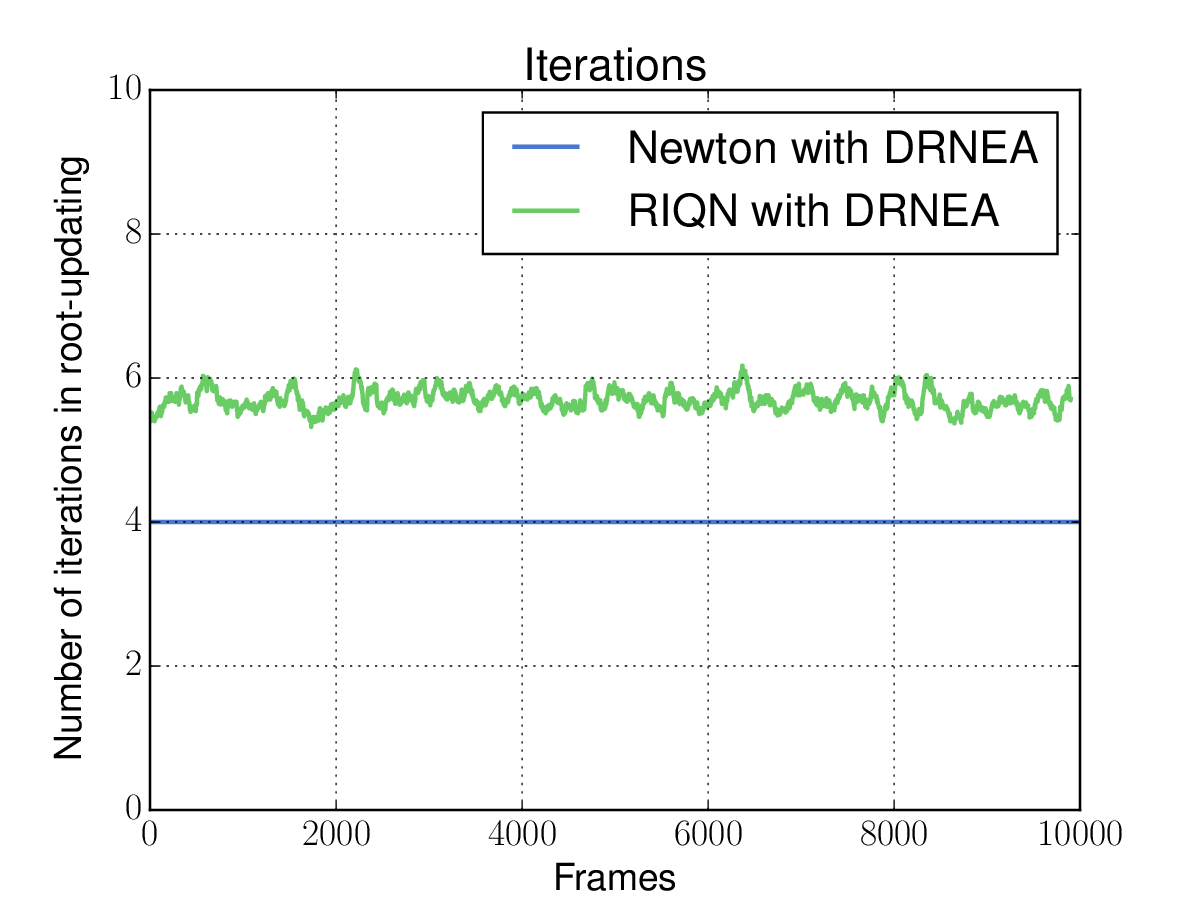}
		\caption{}
		\label{fig:iteration_numbers}
	\end{subfigure}
	\caption{(a) Convergence rate comparison for Newton's method and RIQN 
		(b) Iteration numbers over simulation
                frames. Newton's method: mean = 4, $\sigma =
                0.0$. RIQN: mean = 5.69, $\sigma = 1.16$}
	\label{fig:dummy3}
\end{figure}

We consider the convergent rate of RIQN comparing to Newton’s
method. We inspect the convergence of error
$f(\mathbf{q}_{(l)}^{k+1}) = \mathbf{e}$ during the iterations in
solving the DEL equation for one simulation time step.  For
quantitatively visible convergence, we use the zero configurations as
the initial guess $\mathbf{q}_0^{k+1} = 0$ instead of the proposed initial
guesses in Section \ref{sec:initial_guess}.

Fig. \ref{fig:convergence} shows that under the tolerance RIQN
converges more slowly than Newton's method. This observation is
expected because Newton's method has a quadratic convergence rate which
is in theory faster than that of Quasi-Newton methods. However, in Section 
\ref{sec:performance}, we observed that the absolute computation 
time of the proposed method (DRNEA+RIQN) showed the best performance.

Fig. \ref{fig:iteration_numbers}  shows the average iteration numbers per
each simulation step in the root-finding process. As expected, Newton's 
method requires less iteration numbers than RIQN.

\section{Conclusion}
\label{sec:Conclusion}
We introduced a novel linear-time variational integrator for
simulating multibody dynamic systems. At each simulation time step, the integrator solves a root-finding problem for the DEL
equation using our quasi-Newton algorithm, \emph{RIQN}, which consists of two primary contributions:
\begin{itemize}[label={$\bullet$}]
	\item \textbf{DRNEA:} Based on the variational integrator on Lie group and
	inspired by RNEA, we derived an $O(n)$ recursive algorithm that evaluates DEL
	equations of tree-structured multibody systems. Unlike the previous
	work, which formulates and solves the DEL equation for the entire system,
	in our approach the DEL equation for each body is solved recursively.
	\item \textbf{Root updating: }  By leveraging existing forward
        dynamic algorithm for multibody systems, we introduced
	an $O(n)$ impulse-based dynamic algorithm to estimate the configuration at next time step.
\end{itemize}

We evaluated our linear-time variational integrator on a n-DOF open
chain system and compared the results with existing state-of-art
algorithms. The results show that, for the same computation method of root updating, the performance of our recursive evaluation of the DEL equation (linear-time DEL algorithm) is 15 times faster for a system with 10 degrees of freedom (DOFs) and 32 times faster for 100 DOFs. For the same evaluation method of the DEL equation (\ie linear-time DEL algorithm), our results show that the performance of our new quasi-Newton method is 3.8 times faster for a system with 10 DOFs, and 53 times faster for 100 DOFs. Further analysis shows that for higher DOF systems, the impulse-based Jacobian approximation becomes increasingly more effective compared to our linear-time DEL algorithm.

One of the future directions is to apply the linear-time variational
integrator on constrained dynamic systems. This paper demonstrates the
performance gain on multibody systems with joint constraints, but does
not address other types of constrains, such as contacts or closed-loop
chains. The standard way to handle constraints in a dynamic system is
to solve the DEL equations and constraints simultaneously using
Lagrangian multipliers
\cite{marsden2001discrete,west2004variational}. To preserve the
performance gain achieved by RIQN, one possible extension to
constrained systems is to solve constraint force using the similar idea of
impulse-based forward dynamics 
\cite{featherstone2014rigid,mirtich1995impulse}.

Our current implementation of RIQN can be improved by using variable
time step size. Although the variational integrator allows for larger
time step size than other numerical integrators for the same accuracy,
the variable time step size can still be exploited to achieve further
stability and time performance. However, naively changing the time
step size can have negative impact on the qualitative behavior of a
simulation \cite{hairer2006geometric,kharevych2009geometric}. Previous
work has shown that additional constraints are needed when using
the scheme of variable time step size. Integrating this line of work
to our linear-time variational integrator can be a fruitful future
research direction.

\section*{Acknowledgments}

This work was (partially) funded by the National Science Foundation IIS (\#1409003), Toyota Motor Engineering \& Manufacturing (TEMA), and the Office of Naval Research.

\bibliographystyle{splncs}
\bibliography{dime}
\appendix
\section*{Appendix: Derivative of DRNEA}
\label{app:derivative_of_drnea}

\begin{algorithm}[H]
    \caption{Derivative of DRNEA for computing $\fpp{f(\mathbf{q}^{k+1})}{\mathbf{q}^{k+1}} \in \RE^{n \times n}$}\label{alg:deriv_drnea}
    \begin{algorithmic}[1]
        \For {$j = 1 \to n$}
        \For {$i = 1 \to n$}
        \State {$\fpp{T_{\lambda(i),i}^{k+1}}{q_j^{k+1}} = T_{\lambda(i),i}^{k+1} [S_i] \delta_{ij}$} \Comment{$\delta_{ij} =
        	\begin{cases}
        	1, & \text{if } i = j\\
        	0, & \text{otherwise}
        	\end{cases}
        	$}
        \State {$\fpp{\Delta T_i^k}{q_j^{k+1}} = \left. T_{\lambda(i),i}^{k} \right.^{-1} \fpp{\Delta T_{\lambda(i)}^k}{q_j^{k+1}} T_{\lambda(i),i}^{k+1} + \Delta T_i^{k} [S_i] \delta_{ij}$}
        \State {$\left[ \fpp{V_i^k}{q_j^{k+1}} \right] = \frac{1}{\Delta t} \dlog_{\Delta t V_i^k} \left( \fpp{\Delta T_i^k}{q_j^{k+1}} \exp \left( - \Delta t [V_i^k] \right) \right)$}
        \EndFor
        \For {$i = n \to 1$}
        \State {$\fpp{\mu_i^k}{q_j^{k+1}} = \fpp{}{q_j^{k+1}} \left[ \dlog_{\Delta t V_{i}^{k}} \right]^T G_i V_{i}^{k} + \left[ \dlog_{\Delta t V_{i}^{k}} \right]^T G_i \fpp{V_{i}^{k}}{q_j^{k+1}}$}
        \State {$\fpp{F_i^k}{q_j^{k+1}} = \fpp{\mu_i^k}{q_j^{k+1}} + \sum_{c \in \sigma(i)} \left[ \Ad{\left(T_{i,c}^k\right)^{-1}} \right]^T \fpp{F_{c}^{k}}{q_j^{k+1}} - \fpp{F_i^{\text{ext},k}}{q_j^{k+1}} $}
        \State {$\fpp{f(q^{k+1})}{q_j^{k+1}} = S_i^T \fpp{F_i^k}{q_j^{k+1}} - \fpp{Q_i^k}{q_j^{k+1}}$} \Comment{$j$-th column of $\fpp{f(\mathbf{q}^{k+1})}{\mathbf{q}^{k+1}}$}
        \EndFor
        \EndFor
    \end{algorithmic}
\end{algorithm}


\end{document}